\documentclass[a4paper]{article}

\usepackage{PRIMEarxiv}

\usepackage[utf8]{inputenc} 
\usepackage[T1]{fontenc}    
\usepackage{hyperref}       
\usepackage{url}            
\usepackage{booktabs}       
\usepackage{amsfonts}       
\usepackage{nicefrac}       
\usepackage{microtype}      
\usepackage{lipsum}
\usepackage{fancyhdr}       
\usepackage{graphicx}       
\usepackage{subcaption}
\graphicspath{{media/}}     
\usepackage{multirow}
\usepackage{float}
\usepackage{booktabs}
\usepackage{adjustbox}
\usepackage{placeins}
\hypersetup{
  linktocpage=true,
  colorlinks,
  linkcolor=blue,
  urlcolor=blue,
  citecolor=blue,
  pdfauthor={Asang Dani, Dr. Shailesh R Sathe},
  pdftitle={Cross-Lingual Abstractive Summarization for Marathi},
}

\title{Cross-Lingual Abstractive Summarization for Marathi
}

\author{
  Asang Dani \\
  Department of Computer Science \& Engineering \\
  Visvesvaraya National Institute of Technology \\
  Nagpur, India\\
  \texttt{dt23cse001@students.vnit.ac.in} \\
\And
  Dr. Shailesh R Sathe \\
  Department of Computer Science \& Engineering \\
  Visvesvaraya National Institute of Technology \\
  Nagpur, India\\
  \texttt{srsathe@cse.vnit.ac.in} \\
}

\begin{document}
\maketitle

\begin{abstract}

Text summarization is an important task for advancing natural language
processing (NLP) research and applications in Indic languages. Despite
progress in extractive and abstractive summarization, monolingual
summarization (MS) and cross-lingual summarization (CLS) remain
relatively underexplored for these languages. Recent advances in
Transformer architectures~\cite{transformers-vaswani}, increased
dataset availability, and stronger pretrained language models (PLMs)
offer new opportunities to address this gap. In this work, we
construct cross-lingual training data from the monolingual Marathi
MahaSum corpus~\cite{mahasum-dataset} and combine it with the
substantially smaller cross-lingual PMIndiaSum
dataset~\cite{urlana-pmindiasum} to fine-tune
IndicTrans2~\cite{gumma-etal-2025-towards}, an Indic neural machine
translation (NMT) model. We evaluate the resulting models using ROUGE,
chrF, and chrF++ metrics, focusing primarily on Marathi-to-English CLS
due to computational constraints. We also discuss the applicability of
our approach to the English-to-Marathi direction. Our experiments show
that fine-tuning a pretrained translation model for CLS substantially
outperforms the IndicBART~\cite{dabre_indicbart_2022}- and
mBART-50~\cite{tang2020_mbart_50}-based baselines reported in previous
work on CLS.
  
\end{abstract}


\section{Introduction}
The capabilities of large language models for text-processing tasks
have improved substantially over the past decade. The introduction of
Transformer-based~\cite{transformers-vaswani} deep-learning models was
a major catalyst. Summarization remains one of the most challenging
text-generation tasks and is commonly divided into two categories:
\begin{description}
    \item [Extractive] This approach selects sentences from the source document that convey its most important information.
    \item [Abstractive] This approach generates new text that conveys the source document's key ideas concisely.
\end{description}

Text summarization for English and many European languages has matured
to the point of broad product deployment. In contrast, comparatively
limited work has addressed monolingual and cross-lingual summarization
for Indic languages. Their morphological richness, linguistic
diversity, and limited availability of large high-quality datasets
make the task challenging. This paper investigates abstractive
cross-lingual summarization for the Marathi--English language pair
with focus on Marathi-English direction. It uses a novel approach in
doing CLS. The results vary based on corpus availability and various
other factors.

\section{Related Work}

Before highly capable Indic language models became available, summarization research for Indic languages was constrained by the limited availability of large, high-quality datasets. Apart from the size of the dataset, various other criteria help determine the quality of summarization:

\begin{itemize}

\item Conciseness in the generated summary
\item Completeness in capturing the essence of text
\item Quality of language in generated summary
\item Relevance of summary - does it have unnecessary/irrelevant information
  
\end{itemize}

Zhu et al.~\cite{zhu-etal-2019-ncls} proposed neural cross-lingual
summarization for English--Chinese and Chinese--English. They use
round-trip translation to construct a CLS dataset from a monolingual
summarization corpus and found that neural CLS outperformed
conventional pipeline approaches.

An early study by Sarkar~\cite{sarkar2012bengali} uses an extractive
approach for Bengali. It focuses on monolingual
summarization. HindiSumm~\cite{Singh2024HindiSumm:Dataset} is an
abstractive Hindi summarization corpus comprising 570,000
article--summary pairs from \textit{Navbharat Times} across 21
domains. Authors claim that the dataset has been evaluated against
above criteria by human experts. They fine tuned
mT5~\cite{mt5-xue-2021} and mBART25~\cite{liu-mbart} pre-trained
models on this dataset and evaluate the ensemble performance on this
and various other summarization datasets. They claim that HindiSumm
demonstrates better performance with various datasets, based on
ROUGE-1, ROUGE-2 and ROUGE-N~\cite{lin_rouge_2004} scores.

ILSUM 2.0~\cite{ilsum-20} was created for the
\href{https://ilsum.github.io/ilsum/2024/index.html}{ILSUM shared
  task}. It supports 3 Indian languages - Hindi, Gujarati and
Bengali. For each language approximately 15,000 new-article summary
pairs are included. One of the tasks is monolingual summarization. The
Hindi dataset
\href{https://huggingface.co/datasets/ILSUM/ILSUM-2.0/tree/main/Hindi}
contains 24,225 entries with mean summary length of 39.38 and mean
compression ratio of 14.78. Like Ulrana, IndicBART and
mBART-50~\cite{tang2020_mbart_50} PLMs were fine-tuned and evaluated
using the ROUGE and BERTScore~\cite{zhang_bertscore_2020} metrics.

MahaSum~\cite{mahasum-dataset} is an Marathi summarization dataset
containing approximately 25,000 news articles paired with manually
vetted summaries. Authors fine tuned IndicBART based PLM and claim to
that it is a gold-standard abstractive summarization dataset for
Marathi. They claim significantly better performance better
performance against XL-Sum~\cite{hasan-xlsum} using ROUGE-1, ROUGE-2
and ROUGE-L metrics.

Work by Jothir Adithya T. K. et al.~\cite{t-k-etal-2024-monolingual}
for Marathi, focuses on MS. More recent studies have investigated
cross-lingual abstractive summarization for Indic languages. Deshpande
et al.~\cite{shlok-cross-lingual} proposed two CLS pipelines:
translating the source text before summarization, and summarizing in
the source language before translating the summary. They also proposed
an evaluation metric that combines concept coverage, semantic
similarity, and relevance, and validated it using MuRIL~\cite{muril}.

CrossSum~\cite{bhattacharjee-crosssum} is a cross-lingual dataset that
covers 1500+ language pairs and has around 1.7 million entries. They
fine tuned mT5~\cite{mt5-xue-2021} base model using m2o
(many-to-many), m2m (many-to-many) and o2m (one-to-many) approaches to
evaluate the performance of this dataset. They claim that the MLS
(Multistage Language Sampling) algorithm using m2m approach
consistently outperforms m2o and o2m models. Their evaluation metrics
are ROUGE-2 and their own LaSE (Language-agnostic Summary
Evaluation). Their base models are mT5 and
M2M-100~\cite{fan2020-m2m100}.

Urlana's work~\cite{ulrana-thesis} comprehensively analyses MS and CLS
datasets with respect to summary quality, relevance, readability, and
creativity. He also introduces PMIndiaSum~\cite{urlana-pmindiasum} - a
high-quality multilingual and cross-lingual headline-summarization
dataset covering 14 Indic languages including Marathi. He fine tunes
IndicBART~\cite{dabre_indicbart_2022} and
mBART-50-large~\cite{tang2020_mbart_50} models to evaluate the
performance of this dataset. He also uses ROUGE-2 and ROUGE-L metrics.

After perusing this work, we can clearly see that the choice of
pretrained language model (PLM) strongly influences the quality of MS
and CLS. A model requires robust monolingual/multilingual
understanding for MS/CLS in order to adapt effectively with limited
fine-tuning data. Dataset quality is also quite important. Following
section analyzes various datasets relevant to our work.

\section{Dataset Selection \& Preparation}

Fine-tuning pretrained language models (PLMs) for cross-lingual
summarization requires high-quality document--summary datasets. For
Marathi, available cross-lingual resources are limited in size and
vary in construction quality. We therefore consider four publicly
available resources: MahaSum~\cite{mahasum-dataset},
CrossSum~\cite{bhattacharjee-crosssum}, XL-Sum~\cite{hasan-xlsum}, and
PMIndiaSum~\cite{urlana-pmindiasum}.

These datasets differ in language coverage, source domain, corpus
size, and document--summary construction method, providing
complementary settings for studying Marathi cross-lingual
summarization. MahaSum and the Marathi portion of XL-Sum provide
Marathi monolingual article--summary pairs from which synthetic
cross-lingual training examples are constructed. The English portion
of XL-Sum is also used to construct additional synthetic examples
where appropriate. PMIndiaSum and CrossSum provide cross-lingual
evaluation data involving English and Marathi, although they differ
in source, scale, and data-collection methodology.

We report Cross-lingual summarization results primarily for
Marathi-to-English (MR $\rightarrow$ EN) direction. In the MR
$\rightarrow$ EN direction, the input document is in Marathi and the
target summary is in English.

MahaSum is the primary source of synthetic cross-lingual training
data. For MR $\rightarrow$ EN dataset, we retain the original Marathi
documents and translate the Marathi summaries into English. This
produces Marathi-document/English-summary pairs while preserving the
human-produced Marathi summaries. For EN $\rightarrow$ MR dataset, we
retain the original Marathi summaries and translate the documents into
English. Thus, in both directions, one component of each original
Marathi article--summary pair is machine-translated, while the other
component is retained. As a secondary source of synthetic
cross-lingual dataset, we use XL-Sum corpora.

We use a neural machine translation (NMT) approach for both
translation operations. Machine translation can introduce noise
relative to native cross-lingual data, we evaluate the translation
pipeline separately using round-trip translation. We use
IndicTrans2-200M~\cite{gala-indictrans2} for translation. For
longer-context translation when documents in Marathi are translated to
English, we use the Rotary IndicTrans2 model, which incorporates
rotary positional embeddings (RoPE)~\cite{su_roformer_2024} and
supports longer input sequences~\cite{gumma-etal-2025-towards}. The
translation models also handle code-mixed content appearing in the
MahaSum and XL-Sum corpora.

The resulting dataset configurations and their roles in the
experiments are summarized in Table~\ref{tab:directional_setup}.
The table specifies the document and summary languages, identifies
which field is machine-translated, and distinguishes synthetic
training data from native cross-lingual evaluation data.

\begin{figure*}[htbp]
    \centering
    \includegraphics[
        width=0.95\textwidth
    ]{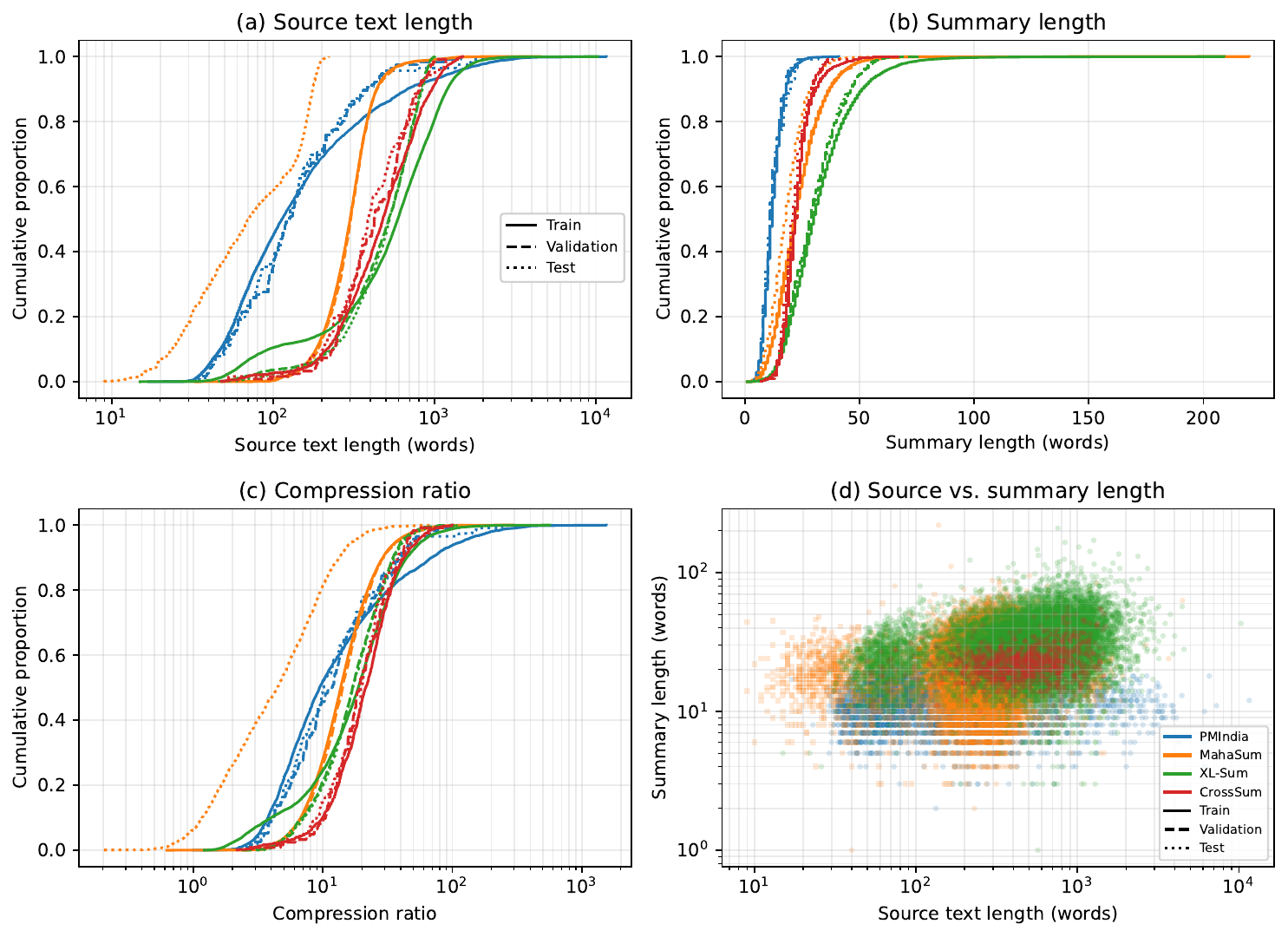}
    \caption{ Distributional comparison of PMIndiaSum, MahaSum, XL-Sum
      and CrossSum.  Panels (a)--(c) show empirical cumulative
      distribution functions (ECDFs) of source-text length, summary
      length, and compression ratio, respectively; panel (d) shows
      source-text versus summary length. Colours denote PMIndiaSum
      (blue), MahaSum (orange), XL-Sum (red) and CrossSum
      (green). Solid and dashed lines denote training and validation
      splits, respectively. }
    \label{fig:dataset-analysis}
\end{figure*}

Table~\ref{tab:dataset-statistics} summarizes the training, validation
and test split statistics for MahaSum, XL-Sum, PMIndiaSum, and
CrossSum. Figure~\ref{fig:dataset-analysis} compares their
document-length, summary-length, and compression-ratio distributions,
as well as the relationship between source and summary lengths.

\paragraph{MahaSum}
MahaSum~\cite{mahasum-dataset} is a Marathi news-summarization corpus
developed to address the limited availability of resources for
abstractive Marathi summarization. The corpus contains 25,374 articles
collected from \textit{Lokmat} and \textit{Loksatta} newspapers; each
instance includes a headline, a summary, and the corresponding article
text. The summaries were manually verified during the corpus
construction.

In our work, MahaSum is the primary source of synthetic CLS training
data. For English-to-Marathi CLS, we retain the original Marathi
summaries and translate the article texts into English. This produces
English-document/Marathi-summary pairs while preserving the
human-produced Marathi summaries. For Marathi-to-English direction, we
retain the original Marathi article texts and translate the summaries
into English. For the translation in both directions, we use an
NMT-based approach. NMT can introduce more noise compared to native
cross-lingual dataset. For NMT, we use
IndicTrans2-200M~\cite{gala-indictrans2}
model. \href{https://huggingface.co/prajdabre/rotary-indictrans2-indic-en-dist-200M}{prajdabre/rotary-indictrans2-indic-en-dist-200M}~\cite{gumma-etal-2025-towards}
is an improved version of original IndicTrans2-200M model as it
supports longer context as it uses Rotary positional embeddings
(RoPE)~\cite{su_roformer_2024}. The translation models can process
code-mixed content appearing in the MahaSum and XL-Sum corpora.

\paragraph{Translation-fidelity validation}

We assess the fidelity of the summary-translation pipeline through round-trip
translation on all 1,667 examples in the MahaSum validation
split. Each original Marathi summary was translated into English and
then back-translated into Marathi. For back-translation, we use
\href{https://huggingface.co/prajdabre/rotary-indictrans2-en-indic-dist-200M}{prajdabre/rotary-indictrans2-en-indic-dist-200M}
model. We compared the resulting Marathi summary with the original
reference using chrF~\cite{popovic-2015-chrf} and
chrF++~\cite{popovic-2017-chrf} metrics. As shown in
Table~\ref{tab:roundtrip-quality}, the pipeline achieved mean chrF and
chrF++ scores of 56.34 and 49.97, respectively. The mean character-
and word-length ratios were 0.967 and 0.994, indicating that summary
length was closely preserved. These results indicate moderate
round-trip lexical fidelity and strong length preservation; they
validate the translation pipeline used to construct the synthetic
data, but do not replace direct evaluation of the final CLS models.

We also assess the fidelity for document/text-translation pipeline
using the round-trip translation approach as described above. As shown
in Table~\ref{tab:roundtrip-text-quality}, the pipeline achieved mean
chrF and chrF++ scores of 63.61 and 59.15, respectively. The mean
character- and word-length ratios were 0.971 and 0.994, indicating
that text length was closely metrics.

Metrics for document translation are marginally better than those
obtained for much shorter summary back-translation. In both the cases,
we manually checked random samples from the synthetic datasets
generated in both the cases and found the translation quality to be
satisfactory.

We didn't evaluate the translation quality for the synthetic XL-Sum
datasets, as the process is computationally expensive. However, given
the above results for MahaSum, the quality is expected to be
good. This is further corroborated by the results obtained after fine
tuning PLMs with these datasets.

\begin{table}[t]
    \centering
    \caption{Round-trip summary translation fidelity on the MahaSum validation split ($N=1{,}667$). Scores compare original Marathi summaries with their Marathi back-translations.}
    \label{tab:roundtrip-quality}
    \begin{tabular}{lrrr}
        \toprule
        \textbf{Metric} & \textbf{Mean} & \textbf{Median} & \textbf{Std. Dev.} \\
        \midrule
        chrF & 56.34 & 56.88 & 11.70 \\
        chrF++ & 49.97 & 50.22 & 11.41 \\
        Character-length ratio & 0.967 & 0.966 & 0.122 \\
        Word-length ratio & 0.994 & 1.000 & 0.348 \\
        \bottomrule
    \end{tabular}
\end{table}

\begin{table}[t]
  \centering
  \caption{Round-trip text translation fidelity on the MahaSum validation split ($N=1{,}667$). Scores compare original Marathi documents with their Marathi back-translations.}
  \label{tab:roundtrip-text-quality}
  \begin{tabular}{lrrr}
    \toprule
    \textbf{Metric} & \textbf{Mean} & \textbf{Median} & \textbf{Std. Dev.} \\
    \midrule
    chrF & 63.01 & 63.61 & 8.17 \\
    chrF++ & 58.69 & 59.15 & 7.71 \\
    Character-length ratio & 0.981 & 0.909 & 0.713 \\
    Word-length ratio & 0.971 & 0.916 & 0.718 \\
    \bottomrule
  \end{tabular}
\end{table}

\paragraph{CrossSum}
CrossSum is a multilingual news-summarization resource constructed
from BBC content and covering 45 languages and 1500+ language
pairs. The complete collection contains approximately 1.7 million
article--summary pairs, including 1,181 Marathi
examples~\cite{bhattacharjee-crosssum}. We use the Marathi subset as
an additional reference corpus alongside PMIndiaSum and to compare
model behaviour across Marathi-English and English-Marathi
summarization resources. Its the smallest of CLS datasets we use in
our work. As noted by the authors, dataset is not curated by humans.

\paragraph{XL-Sum}
XL-Sum is a multilingual news-summarization resource constructed from
BBC content and covering 44 languages. We use the Marathi corpus as
additional Marathi-English summarization resource by translating
summary field to English. Similarly, we use the English subset which
which is orders of magnitude larger as an additional resource for
English-Marathi CLS by translating the summary field to
Marathi. reference corpus and to compare model behaviour across
Marathi-English and English-Marathi summarization resources. Like
MahaSum, we use translation approach to construct synthetic CLS
data-set from XL-Sum-MR and XL-Sum-EN datasets.

\paragraph{PMIndiaSum}
PMIndiaSum is a multilingual and cross-lingual summarization resource
for Indian languages. It support 14 languages and 196 language
directions. The dataset provides both monolingual and cross-lingual
document--headline pairs~\cite{urlana-pmindiasum}. It contains
approximately 4,400 examples. All texts have been translated by humans
as they belong to the official press releases by India PM office. We
use this as primary reference in our work. Its is too small to be
useful as the only dataset. As a result, we use it in combination with
XL-Sum and MahaSum. We split it in 3 parts - train, validation and
test split. The \emph{test} split is never seen during fine-tuning. It
is used to evaluate the quality of fine-tuned models.

\begin{table*}[t]
\centering
\caption{Marathi-English dataset statistics for the training, validation,
and test splits. Lengths are measured in words, and compression ratio is
defined as source-text length divided by summary length.}
\label{tab:dataset-statistics}
\small
\begin{tabular}{llrrrrrrrrrr}
\toprule
Dataset & Split & $N$ &
\multicolumn{3}{c}{Text length} &
\multicolumn{3}{c}{Summary length} &
\multicolumn{3}{c}{Compression} \\
\cmidrule(lr){4-6}
\cmidrule(lr){7-9}
\cmidrule(lr){10-12}
&&& Mean & Median & P95 & Mean & Median & P95
& Mean & Median & P95 \\
\midrule

MahaSum & Train &
22000 & 316.99 & 300 & 511.00 &
23.32 & 22 & 42.0 &
16.67 & 13.75 & 36.94 \\

MahaSum & Val &
1667 & 320.70 & 302 & 528.10 &
23.10 & 22 & 42.0 &
17.08 & 14.20 & 37.64 \\

MahaSum & Test &
1667 & 91.03 & 71 & 187.00 &
18.48 & 18 & 32.0 &
6.21 & 4.47 & 17.25 \\

\midrule

XL-Sum & Train &
10903 & 653.28 & 592 & 1424.90 &
32.57 & 30 & 60.0 &
23.34 & 18.76 & 58.15 \\

XL-Sum & Val &
1362 & 519.98 & 521 & 877.00 &
29.87 & 29 & 51.0 &
19.58 & 16.94 & 40.29 \\

XL-Sum & Test &
1362 & 529.32 & 528 & 899.00 &
30.13 & 29 & 51.0 &
20.10 & 16.97 & 42.43 \\

\midrule

PMIndia & Train &
4179 & 298.63 & 113 & 1292.10 &
12.31 & 12 & 20.0 &
28.56 & 9.42 & 120.27 \\

PMIndia & Val &
116 & 187.93 & 127 & 473.25 &
11.91 & 11 & 19.0 &
17.28 & 11.62 & 48.17 \\

PMIndia & Test &
116 & 211.77 & 126 & 489.00 &
12.36 & 11 & 22.0 &
19.70 & 10.08 & 47.89 \\

\midrule

CrossSum & Train &
899 & 537.33 & 479 & 1102.20 &
23.00 & 22 & 36.0 &
24.40 & 21.43 & 53.40 \\

CrossSum & Val &
123 & 477.91 & 423 & 913.10 &
22.38 & 22 & 31.9 &
22.58 & 19.32 & 44.53 \\

CrossSum & Test &
123 & 467.18 & 392 & 962.50 &
21.69 & 21 & 31.9 &
22.34 & 20.12 & 50.80 \\

\bottomrule
\end{tabular}
\end{table*}

\begin{table}[t]
\centering
\caption{Construction and use of datasets for the two cross-lingual summarization 
directions. ``Translated field'' identifies the field generated using IndicTrans2.}
\label{tab:directional_setup}
\small
\begin{tabular}{lllllll}
\toprule
Direction & Dataset & Original document & Original summary
& Translated field & Data type & Role \\
\midrule
MR $\rightarrow$ EN
& MahaSum
& MR
& MR
& SUMM
& Synthetic
& Training \\

MR $\rightarrow$ EN
& XL-Sum-MR
& MR
& MR
& Summary
& Synthetic
& Training \\

MR $\rightarrow$ EN
& PMIndiaSum
& EN
& MR
& None
& Native
& Evaluation \\

MR $\rightarrow$ EN
& CrossSum
& EN
& MR
& None
& Native
& Evaluation \\

\midrule

EN $\rightarrow$ MR
& MahaSum
& MR
& MR
& TEXT
& Synthetic
& Training \\

EN $\rightarrow$ MR
& XL-Sum-EN
& EN
& EN
& Summary
& Synthetic
& Training \\

EN $\rightarrow$ MR
& PMIndiaSum
& EN
& MR
& None
& Native
& Evaluation \\

EN $\rightarrow$ MR
& CrossSum
& EN
& MR
& None
& Native
& Evaluation \\

\bottomrule
\end{tabular}
\end{table}

\section{Summarization Approach and PLM Selection}

Our goal is to achieve high quality cross-lingual summarization for
Marathi. We considered several PLMs for fine-tuning. The principal
criterion was strong coverage of Marathi, other Indic languages, and
English. Although BART-based models~\cite{bart-lewis-2020} are
effective for abstractive MS, adapting them to CLS can be
challenging. We therefore investigate a novel approach based on
IndicTrans2~\cite{gala-indictrans2} PLM which supports all 22
scheduled Indian languages. We investigate the use of IndicTrans2 as
the primary generative backbone for Marathi–English cross-lingual
summarization. IndicTrans2 is known to have excellent cross-lingual
understanding and performs at par with or better than most other Open
Source models NLLB~\cite{2022_nllb} and M2M-100~\cite{fan2020-m2m100},
and commercial ones from Microsoft and Google. It performs well in our
experiments for both English-to-Marathi and Marathi-to-English CLS.

We use FlashAttention-2~\cite{dao_flashattention-2_2023} during
fine-tuning to improve training speed and reduce memory
consumption. The primary models examined for Marathi-English CLS were:

\begin{enumerate}
  \item \texttt{indictrans2-indic-en-1B}
  \item \texttt{rotary-indictrans2-indic-en-1B}
  \item \texttt{indictrans2-indic-en-dist-200M}
  \item \texttt{rotary-indictrans2-indic-en-dist-200M}
\end{enumerate}

We also study Rotary positional embeddings
(RoPE)~\cite{su_roformer_2024} based models as they enable contexts of
up to 2,048 tokens. For English-Marathi direction, corresponding
\texttt{en-indic} models were used. In contrast to the work of Urlana
et al.~\cite{urlana-pmindiasum}, we do not use
IndicBART~\cite{dabre_indicbart_2022} or mT5~\cite{mt5-xue-2021}.

\section{Experimental Setup}

Our training pipeline uses the
\href{https://huggingface.co/learn/llm-course/chapter3/3?fw=pt}{Hugging
  Face Trainer API}. We fine-tune the PLMs on NVIDIA A100 GPUs with
40GB VRAM\footnote{This work was generously supported by
\href{https://cdac.in/}{C-DAC} through access to the
\href{https://cdac.in/index.aspx?id=hpc_nsf_siddhi-AI}{C-DAC PARAM
  Siddhi AI} HPC cluster.}. The pipeline uses \texttt{torchrun}, a
PyTorch~\cite{paszke2019pytorch} utility, to distribute fine-tuning
across multiple GPUs. For repeatability, accuracy and reproducibility,
we used MLflow~\cite{zaharia-mlflow} to log training parameters, track
metrics, and preserve artifacts for each run. Checkpointing was
enabled to enable resuming the training run after interruption, and
the best-performing checkpoint from each run was retained.

The experiments are organized by the two directions defined in
Table~\ref{tab:directional_setup}. All reported results are therefore
identified explicitly as either Marathi-to-English. However we have
reported the numbers for Marathi-to-English direction as this
direction is more challenging and significant GPU resources are needed
to obtain identical number in the English-Marathi direction.

\begin{table}[H]
    \centering
    \caption{Hyperparameters considered for fine-tuning the IndicTrans2 model}
    \label{tab:indictrans2_hparams}
    \begin{tabular}{llp{5.8cm}}
    \hline
    \textbf{Hyperparameter} & \textbf{Value / Range} & \textbf{Description} \\
    \hline
    Learning rate & 1e-5 -- 5e-5 & Range explored; lower values can reduce catastrophic forgetting on smaller datasets \\
    Batch size & 2 -- 8 & Batch-size range considered; gradient accumulation may be used to increase the effective batch size \\
    Optimizer & Adam ($\beta_1=0.9$, $\beta_2=0.98$, $\epsilon=10^{-8}$) & Standard optimizer for transformer models \\
    Learning rate scheduler & Inverse square root / linear decay & Schedulers considered during fine-tuning \\
    Dropout & 0.1 -- 0.3 & Regularisation range explored during fine-tuning \\
    Maximum sequence length & 256 -- 2048 & Maximum number of tokens in each source--target pair \\
    Epochs & 3 -- 5 & Selected according to dataset size and early stopping \\
    Seed & 42 & Used for reproducibility \\
    Evaluation metrics & ROUGE-1, ROUGE-2, ROUGE-L, chrF++ & ROUGE-L was used for checkpoint selection \\
    \hline
    \end{tabular}
\end{table}

Based on our experiments, we found following parameters to be most optimal:

\begin{itemize}
\item learning rate - $4e^-5$
\item Dropout rate - $0.1$
\item Optimzer - $adam\_w$
\item Batch Size - $8$
\item Epochs - $3$
\end{itemize}

\section{Evaluation and Validation}

ROUGE~\cite{lin_rouge_2004} is a widely used metric for text
summarization, and BERTScore~\cite{zhang_bertscore_2020} is commonly
used to evaluate generated text semantically. We use ROUGE-L to select
the best-performing checkpoint from each training run. We also
manually inspect generated summaries for meaningfulness, coherence,
and coverage of the source document's central content.

\section{Results}

We first establish the baseline performance on the PMIndia-MR test set
using IndicBART~\cite{dabre_indicbart_2022} and
mBART-50~\cite{tang2020_mbart_50} models. The numbers are not readily
available elsewhere in published literature. We fine-tune the baseline
models with same inputs as our IndicTrans2 models. For the baseline,
ROUGE and BERTScore-F1 metrics are reported in
Table~\ref{tab:indicbart-baseline} and
Table~\ref{tab:mbart50-baseline} report the baseline numbers. Best
results are obtain across all the metrics for baseline model
fine-tuned with PMIndiaSum+MahaSum dataset. It also achieves highest
chrF and chrF++ scores at 31.8064 \& 30.2792, respectively.

\begin{figure}[htbp]
  \centering
    \begin{subfigure}[b]{0.47\linewidth}
        \centering
        \includegraphics[width=\linewidth]{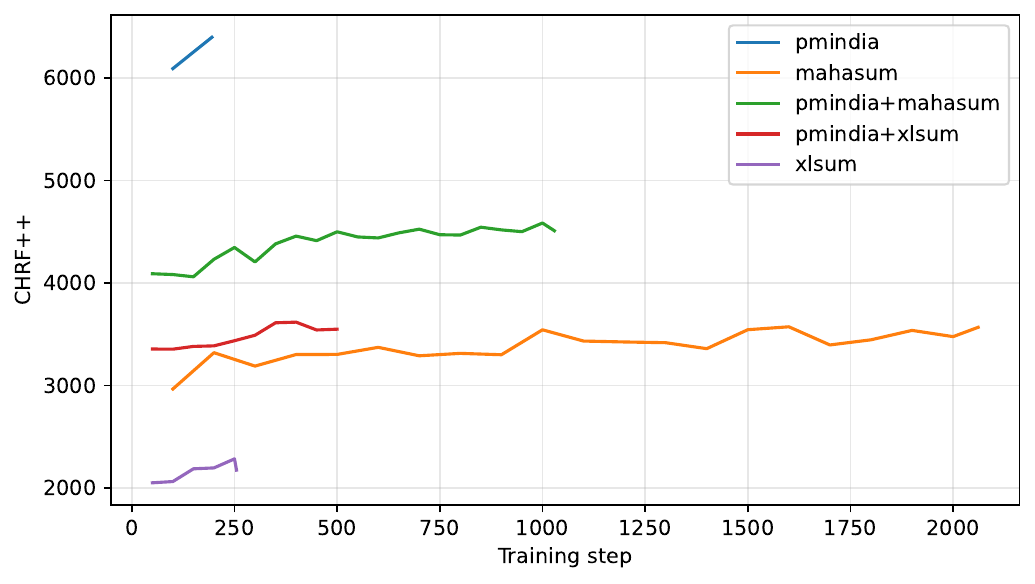}
        \caption{chrF++}
        \label{fig:eval-chrfpp}
    \end{subfigure}
    \hfill
    \begin{subfigure}[b]{0.47\linewidth}
        \centering
        \includegraphics[width=\linewidth]{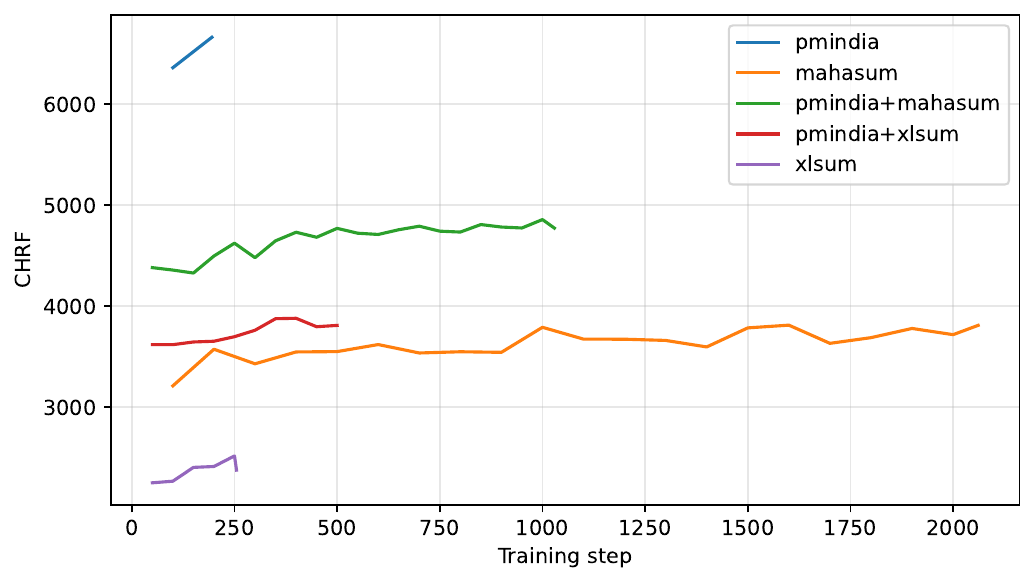}
        \caption{chrF}
        \label{fig:eval-chrf}
    \end{subfigure}
    \\[0.8em]
    \begin{subfigure}[b]{0.47\linewidth}
        \centering
        \includegraphics[width=\linewidth]{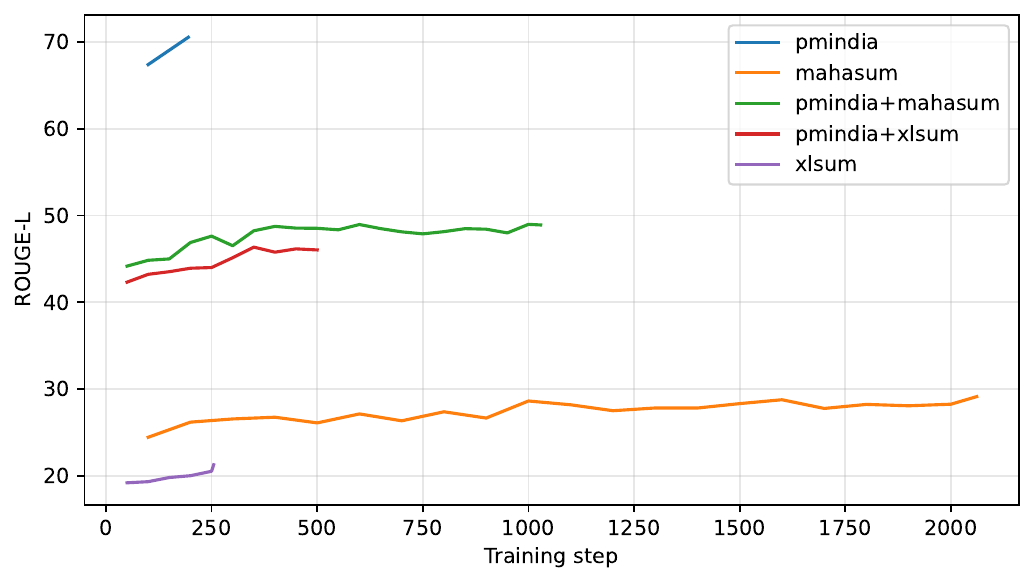}
        \caption{ROUGE-L}
        \label{fig:eval-rougeL}
    \end{subfigure}
    \hfill
    \begin{subfigure}[b]{0.47\linewidth}
        \centering
        \includegraphics[width=\linewidth]{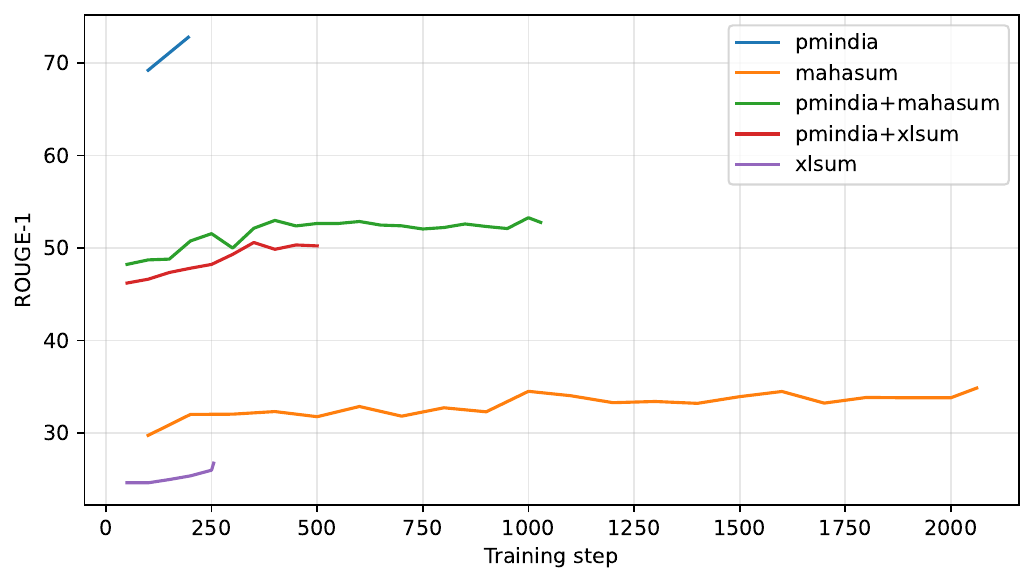}
        \caption{ROUGE-1}
        \label{fig:eval-rouge1}
    \end{subfigure}
    \\[0.8em]
    \begin{subfigure}[b]{0.47\linewidth}
        \centering
        \includegraphics[width=\linewidth]{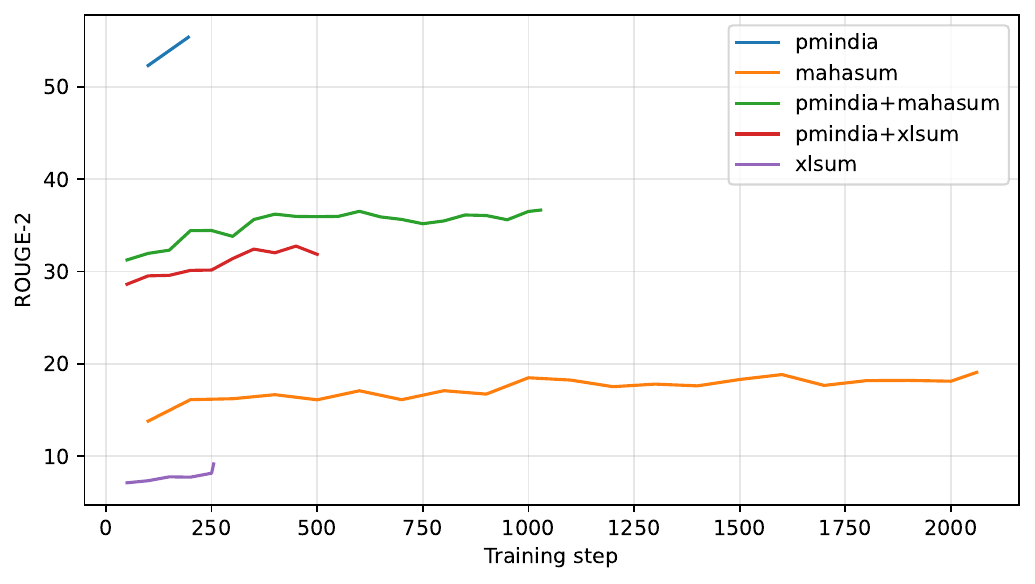}
        \caption{ROUGE-2}
        \label{fig:eval-rouge2}
    \end{subfigure}
    \caption{Validation-metric trajectories during Marathi-to-English
      fine-tuning of \texttt{indictrans2-200M} model. Each line
      corresponds to training on PMIndiaSum, MahaSum, XL-Sum and their
      blended training set, as identified in the in-panel legends. The
      panels are enlarged to make the training dynamics and dataset
      labels legible; final held-out test-set scores should be
      reported separately in a results table.}
    \label{fig:comp-mahasum-xlsum}
\end{figure}

Figure~\ref{fig:comp-mahasum-xlsum} shows training and validation
metric trajectories for \texttt{indictrans2-indic-en-dist-200M}
fine-tuning.

Table~\ref{tab:mren_data_ablation} presents a training-data ablation
for MR$\rightarrow$EN cross-lingual summarization. All models were
initialized from the \texttt{indictrans2-indic-en-dist-200M}
checkpoint and evaluated on the same held-out PMIndiaSum
MR$\rightarrow$EN test set.  The model trained on the combined
PMIndiaSum and MahaSum datasets achieved the strongest performance,
with ROUGE-1, ROUGE-2, and ROUGE-L scores of 75.65, 56.37, and 72.58,
respectively, as well as a BERTScore-F1 of 99.15. Compared with
PMIndiaSum-only training, the combined-data configuration improved
ROUGE-1, ROUGE-2, and ROUGE-L by 3.11, 4.02, and 3.17 points,
respectively.

Adding XL-Sum Marathi to PMIndiaSum resulted in smaller improvements,
increasing ROUGE-1 from 72.55 to 72.99, ROUGE-2 from 52.36 to 53.72,
and ROUGE-L from 69.41 to 70.35. In contrast, models trained
exclusively on MahaSum or XL-Sum Marathi performed substantially worse
than the models trained on PMIndiaSum. The MahaSum-only configuration
obtained ROUGE-1, ROUGE-2, and ROUGE-L scores of 40.96, 23.23, and
37.42, respectively, whereas the XL-Sum-Marathi-only configuration
obtained scores of 40.35, 23.64, and 37.22. These differences may
reflect variation in dataset size, domain, summary style,
preprocessing, or the alignment between each dataset and the
PMIndiaSum evaluation distribution.

\begin{figure}[htbp]
  \centering
    \begin{subfigure}[b]{0.47\linewidth}
        \centering
        \includegraphics[width=\linewidth]{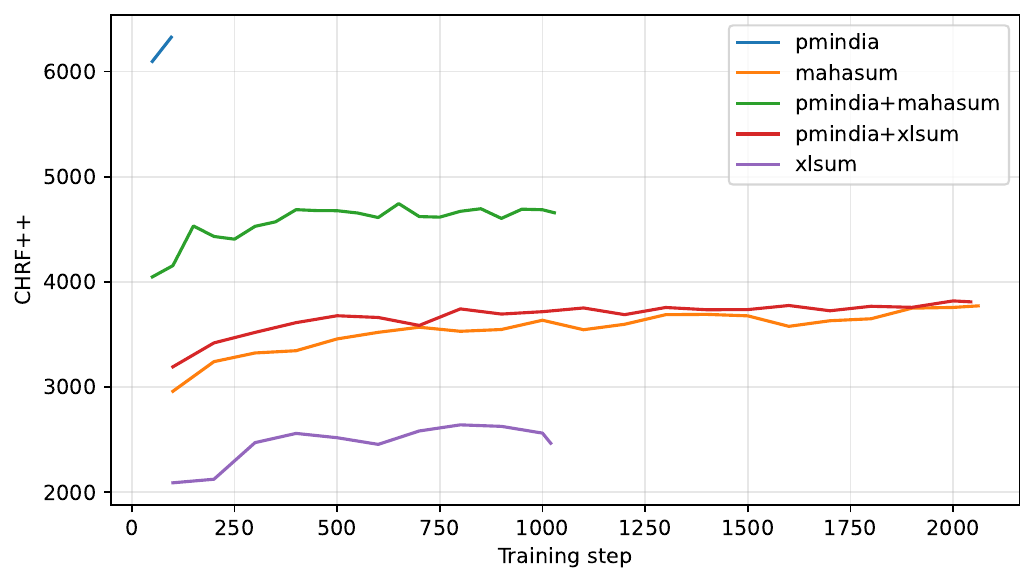}
        \caption{chrF++ - IndicTrans 1B}
        \label{fig:eval-chrfpp}
    \end{subfigure}
    \hfill
    \begin{subfigure}[b]{0.47\linewidth}
        \centering
        \includegraphics[width=\linewidth]{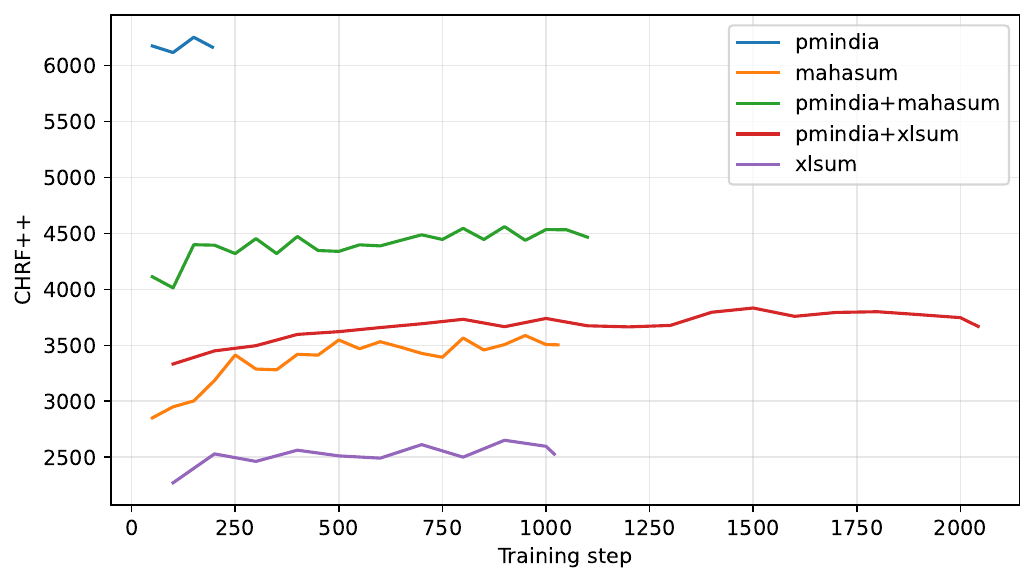}
        \caption{chrF++ - Rotary IndicTrans 1B}
        \label{fig:eval-chrfpp-rot-1b}
    \end{subfigure}
    \\[0.8em]
    \begin{subfigure}[b]{0.47\linewidth}
        \centering
        \includegraphics[width=\linewidth]{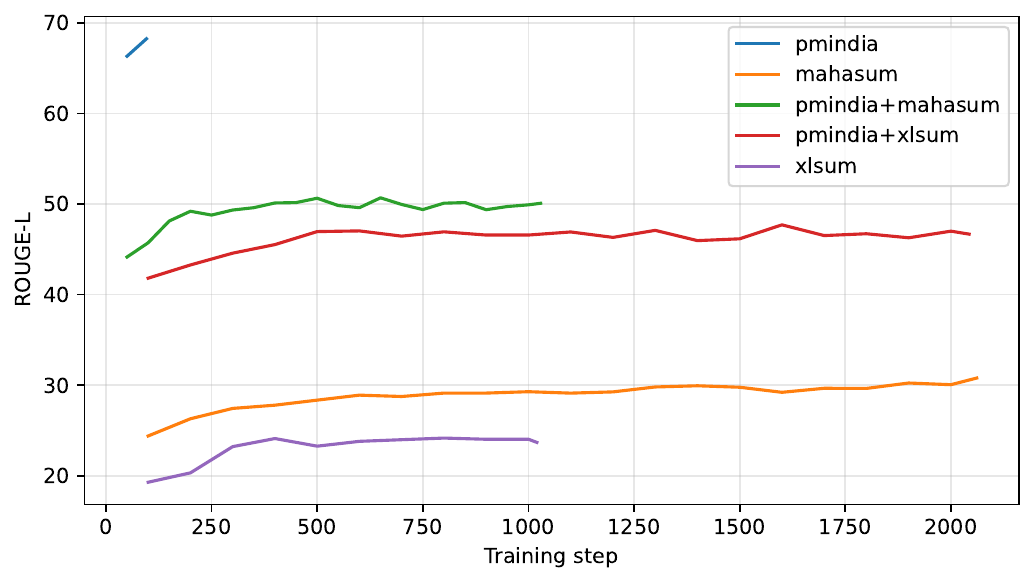}
        \caption{ROUGE-L-IndicTrans 1B}
        \label{fig:eval-rougeL-1b}
    \end{subfigure}
    \hfill
    \begin{subfigure}[b]{0.47\linewidth}
        \centering
        \includegraphics[width=\linewidth]{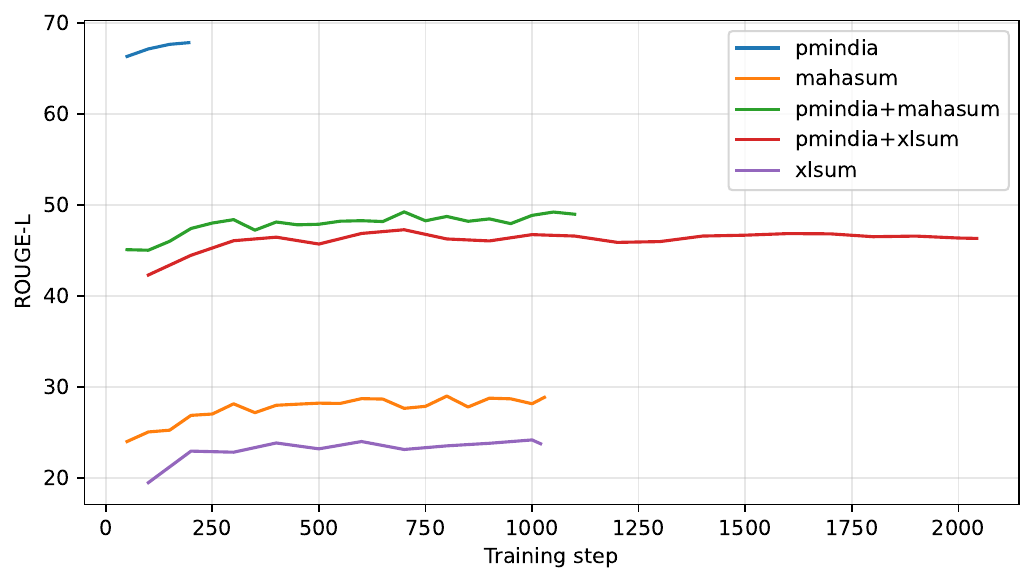}
        \caption{ROUGE-L- Rotary IndicTrans 1B}
        \label{fig:eval-rougeL-rot-1b}
    \end{subfigure}
    \caption{Validation-metric trajectories during Marathi-to-English
      fine-tuning of 1B parameter models.}
    \label{fig:comp-mahasum-xlsum-1b}
\end{figure}

We also evaluate 1 billion parameter
model. Figure~\ref{fig:comp-mahasum-xlsum-1b} shows training and
validation trajectories for \texttt{indictrans2-indic-en-1B} fine
tuning.

Table~\ref{tab:mren_rouge_ablation_1b} presents a training-data
ablation for MR$\rightarrow$EN cross-lingual summarization using
\texttt{indictrans2-indic-en-1B}. All models were evaluated on the
same held-out PMIndiaSum MR$\rightarrow$EN test set. The model
trained on the combined PMIndiaSum and MahaSum datasets achieved the
strongest ROUGE performance, obtaining ROUGE-1, ROUGE-2, and ROUGE-L
scores of 72.50, 53.31, and 69.81, respectively. Compared with
PMIndiaSum-only training, this configuration improved ROUGE-1 by 0.48
points, ROUGE-2 by 0.11 points, and ROUGE-L by 0.10 points. It also
achieved the highest chrF and chrF++ scores, at 64.99 and 62.57,
respectively.

Adding XL-Sum Marathi to PMIndiaSum produced a mixed effect: ROUGE-1
increased from 72.02 to 72.25, whereas ROUGE-2 decreased from 53.20 to
52.72 and ROUGE-L decreased from 69.71 to 69.45. Models trained
exclusively on MahaSum or XL-Sum Marathi performed substantially worse
than those trained with PMIndiaSum. These results suggest that MahaSum
provides limited but measurable complementary information when combined
with PMIndiaSum, whereas XL-Sum Marathi does not consistently improve
performance under the present experimental setup.

\begin{table*}[t]
    \centering
    \caption{IndicBART Baseline MR$\rightarrow$EN cross-lingual summarization
      results on the PMIndia-MR test set for different training-data
      configurations.}
    \label{tab:indicbart-baseline}
    \begin{adjustbox}{max width=\textwidth}
    \begin{tabular}{llrrrrrrrr}
        \toprule
        Base model
        & Training data
        & ROUGE-1
        & ROUGE-2
        & ROUGE-L
        & BERTScore-F1 \\
        \midrule
        Baseline run
        & PMIndia-MR
        & 36.6299 & 17.9590 & 33.7923 & 98.1233 \\

        Baseline run
        & PMIndia-MR + MahaSum-MR
        & \textbf{40.2094} & \textbf{20.1573} &
        \textbf{37.7508} & \textbf{98.3127} \\

        Baseline run
        & PMIndia-MR + CrossSum-MR
        & 19.0129 & 5.1334 & 17.3480 & 97.6011 \\

        Baseline run
        & PMIndia-MR + XLSum-Marathi
        & 28.0603 & 12.3473 & 26.2806 & 97.9843 \\
        \bottomrule
    \end{tabular}
    \end{adjustbox}
\end{table*}

\begin{table*}[t]
    \centering
    \caption{mBART-50 Baseline MR$\rightarrow$EN cross-lingual summarization
      results on the PMIndia-MR test set for different training-data
      configurations.}
    \label{tab:mbart50-baseline}
    \begin{adjustbox}{max width=\textwidth}
    \begin{tabular}{llrrrrrrrr}
        \toprule
        Base model
        & Training data
        & ROUGE-1
        & ROUGE-2
        & ROUGE-L
        & BERTScore-F1 \\
        \midrule
        Baseline run
        & PMIndia-MR
        & 60.46 & 39.435 & 58.596 & 98.8282 \\

        Baseline run
        & PMIndia-MR + MahaSum-MR
        & \textbf{62.0834} & \textbf{39.7034} & \textbf{59.7904} & \textbf{98.8772} \\

        Baseline run
        & PMIndia-MR + CrossSum-MR
        & 59.8894 & 38.3986 & 57.5647 & 98.8161 \\

        Baseline run
        & PMIndia-MR + XLSum-Marathi
        & 59.1948 & 37.3118 & 57.3575 & 98.8161 \\
        \bottomrule
    \end{tabular}
    \end{adjustbox}
\end{table*}

\begin{table*}[t]
    \centering
    \caption{MR$\rightarrow$EN cross-lingual summarization results for
      IndicTrans2 fine-tuned with different training-data
      configurations.  The base model for all experiments is
      \texttt{indictrans2-indic-en-dist-200M}. All systems are
      evaluated on the held-out PMIndiaSum MR$\rightarrow$EN test
      set.}
    \label{tab:mren_data_ablation}
    \begin{adjustbox}{max width=\textwidth}
    \begin{tabular}{llrrrr}
        \toprule
        Base model
        & Training data
        & ROUGE-1
        & ROUGE-2
        & ROUGE-L
        & BERTScore-F1 \\
        \midrule
        IndicTrans2-dist-200M
        & PMIndiaSum
        & 72.5478 & 52.3571 & 69.4082 & 99.0716 \\

        IndicTrans2-dist-200M
        & MahaSum
        & 40.9599 & 23.2271 & 37.4243 & 98.6202 \\

        IndicTrans2-dist-200M
        & PMIndiaSum + MahaSum
        & \textbf{75.6536}
        & \textbf{56.3743}
        & \textbf{72.5812}
        & \textbf{99.1497} \\

        IndicTrans2-dist-200M
        & PMIndiaSum + XL-Sum Marathi
        & 72.9930 & 53.7196 & 70.3513 & 99.0991 \\

        IndicTrans2-dist-200M
        & XL-Sum Marathi
        & 40.3488 & 23.6426 & 37.2245 & 98.5627 \\
        \bottomrule
    \end{tabular}
    \end{adjustbox}
\end{table*}

\begin{table}[t]
    \centering
    \caption{MR$\rightarrow$EN cross-lingual summarization results for
    the IndicTrans2 1B model under different training-data
    configurations.}
    \label{tab:mren_rouge_ablation_1b}
    \begin{tabular}{lrrrr}
        \toprule
        Training data
        & ROUGE-1
        & ROUGE-2
        & ROUGE-L
        & BERTScore-F1 \\
        \midrule
        PMIndiaSum
        & 72.0151 & 53.2036 & 69.7104 & 99.0504 \\

        MahaSum
        & 40.4998 & 22.3857 & 36.7866 & 98.5998 \\

        PMIndiaSum + MahaSum
        & \textbf{72.4968}
        & \textbf{53.3091}
        & \textbf{69.8116}
        & 99.0485 \\

        PMIndiaSum + XL-Sum Marathi
        & 72.2468 & 52.7246 & 69.4549 & 99.0442 \\

        XL-Sum Marathi
        & 37.5122 & 20.8242 & 34.0955 & 98.5330 \\
        \bottomrule
    \end{tabular}
\end{table}

\begin{figure}[htbp]
  \centering
    \begin{subfigure}[b]{0.47\linewidth}
        \centering
        \includegraphics[width=\linewidth]{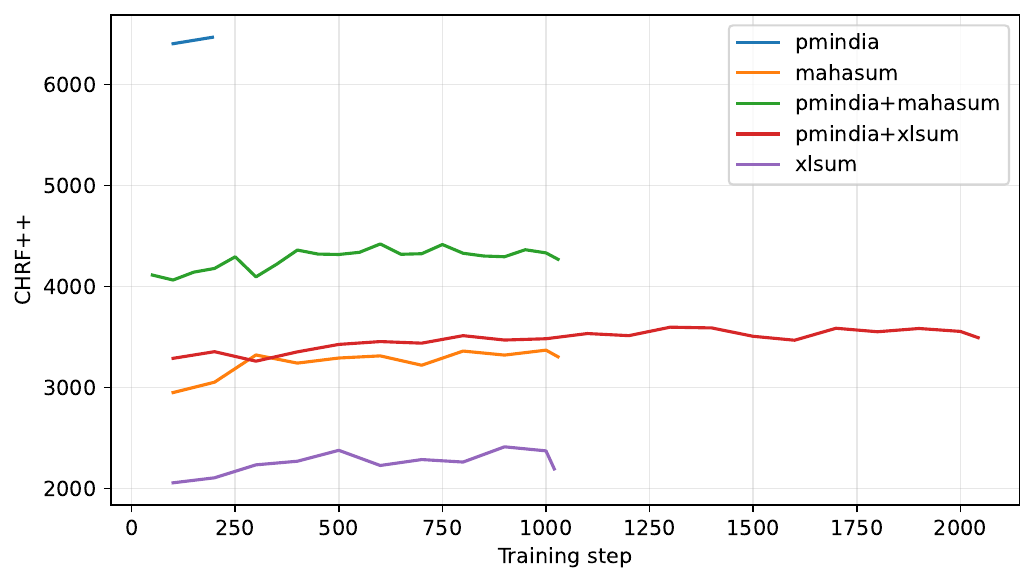}
        \caption{chrF++}
        \label{fig:eval-chrfpp}
    \end{subfigure}
    \hfill
    \begin{subfigure}[b]{0.47\linewidth}
        \centering
        \includegraphics[width=\linewidth]{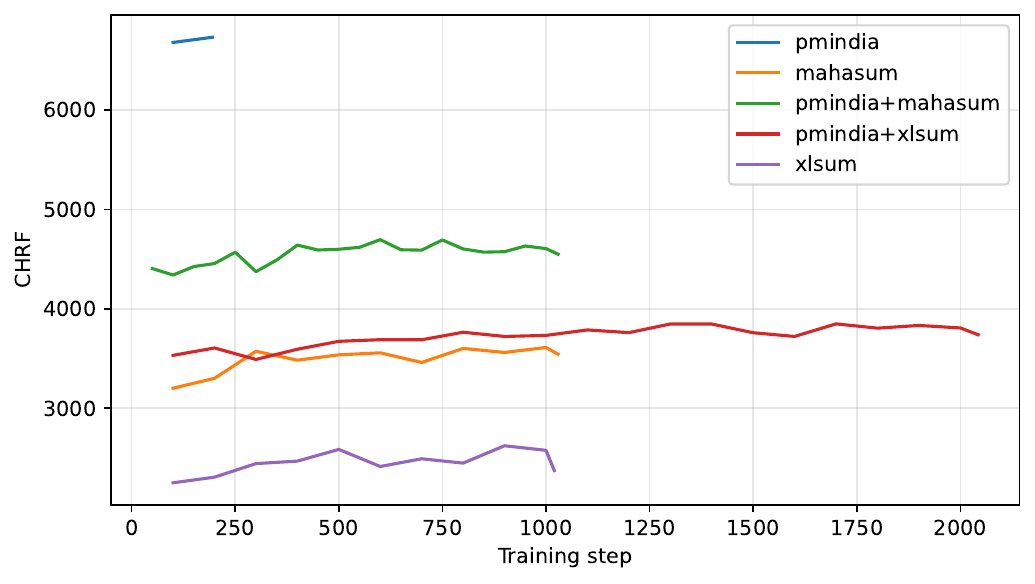}
        \caption{chrF}
        \label{fig:eval-chrf}
    \end{subfigure}
    \\[0.8em]
    \begin{subfigure}[b]{0.47\linewidth}
        \centering
        \includegraphics[width=\linewidth]{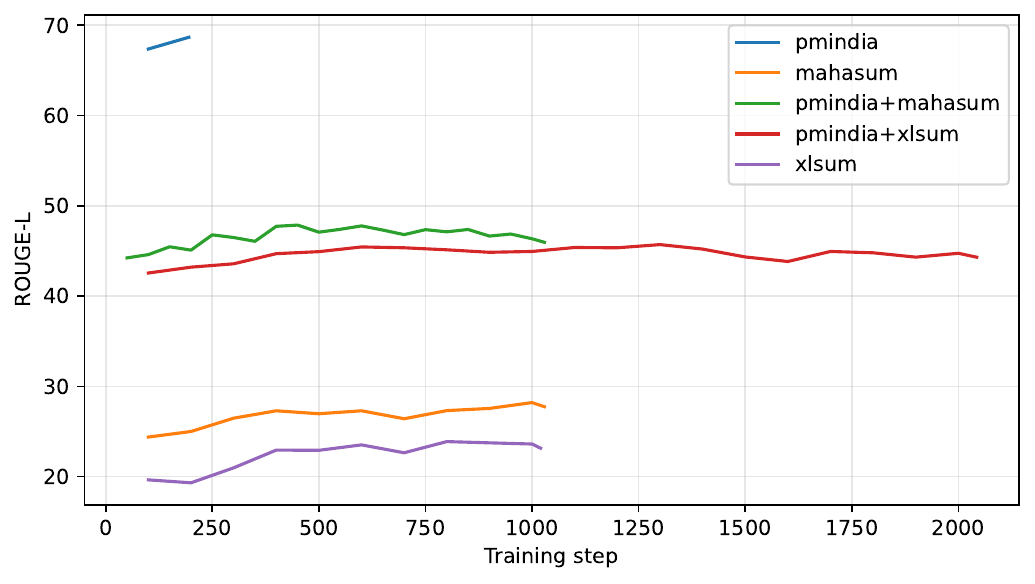}
        \caption{ROUGE-L}
        \label{fig:eval-rougeL}
    \end{subfigure}
    \hfill
    \begin{subfigure}[b]{0.47\linewidth}
        \centering
        \includegraphics[width=\linewidth]{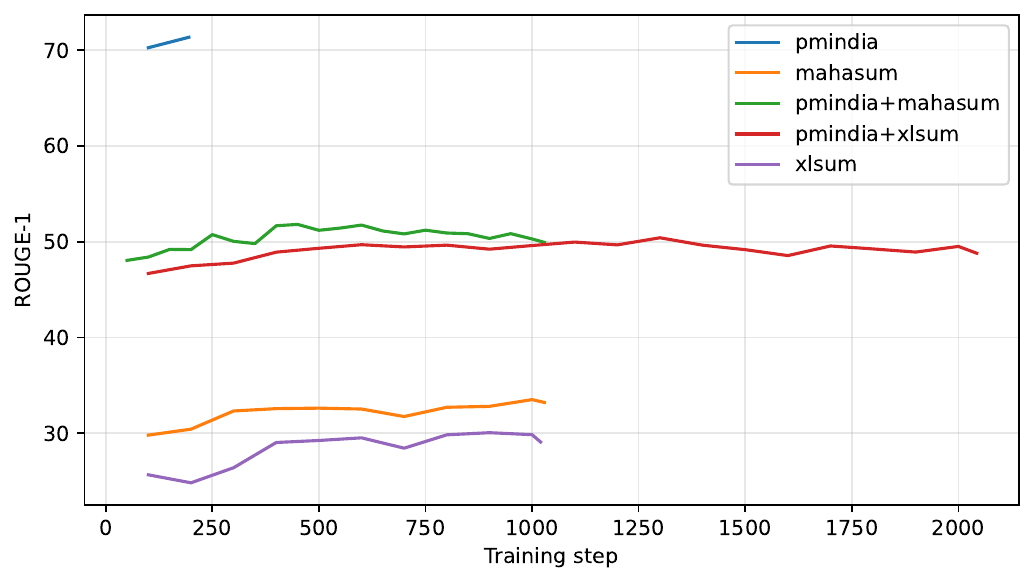}
        \caption{ROUGE-1}
        \label{fig:eval-rouge1}
    \end{subfigure}
    \\[0.8em]
    \begin{subfigure}[b]{0.47\linewidth}
        \centering
        \includegraphics[width=\linewidth]{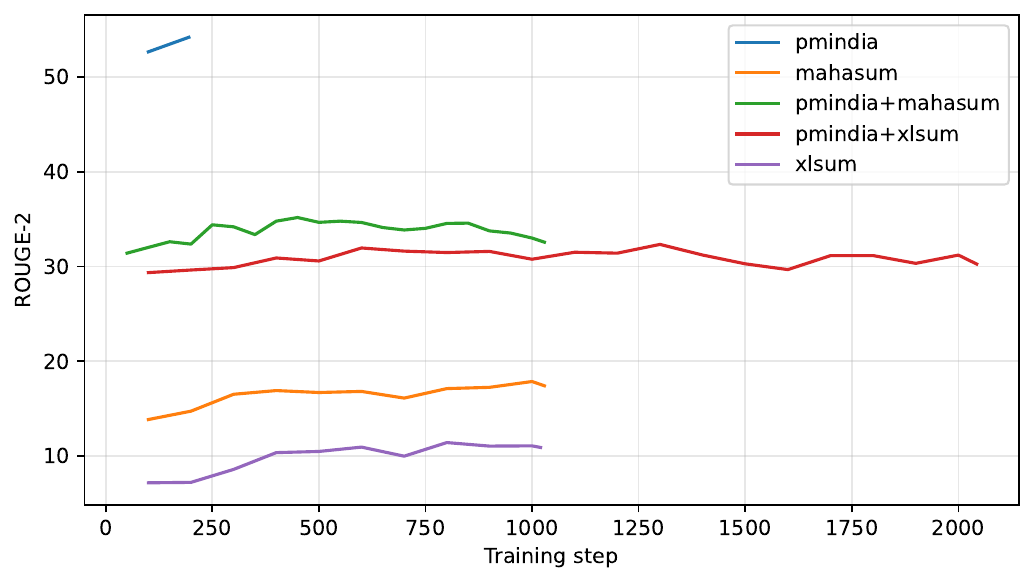}
        \caption{ROUGE-2}
        \label{fig:eval-rouge2}
    \end{subfigure}
    \caption{Validation-metric trajectories during Marathi-to-English
      fine-tuning of \texttt{rotary-indictrans2-200M} model. Each line
      corresponds to training on PMIndiaSum, MahaSum, XL-Sum and their
      blended training set, as identified in the in-panel legends. The
      panels are enlarged to make the training dynamics and dataset
      labels legible; final held-out test-set scores should be
      reported separately in a results table.}
    \label{fig:comp-mahasum-xlsum-rot-it2-200m}
\end{figure}

Based on these results, we found that best performing model (SOTA) for
Marathi-English CLS is the IndicTrans2-200M PLM fine tuned on
PMIndiaSum+MahaSum dataset. Our approach improves upon the existing
baseline significantly using the most important datasets. Why smaller
200M models performs better than 1B model needs further
investigation.

\section{Conclusion}

Cross-lingual summarization remains challenging for Indian
languages. Continued advances in pretrained language models and the
availability of larger, higher-quality datasets should improve
performance. Future work should investigate stronger robust human
evaluation, and methods for improving factual consistency and
cross-lingual faithfulness. Lowering inference cost is also important
as it makes multi-lingual content much more accessible.

\section{Future Work}
We plan to explore how much multilingual capacity is actually
necessary for cross-lingual summarization. Exploring Deep
Compression~\cite{han_deep_2016} approaches to reduce the inference
cost of Indic NLP models is the logical next step. IndicTrans2 shares
decoder input/output embeddings, and the decoder output projection
represents a substantial portion of the parameter count. Our goal is
to focus on English-Marathi language pair. We hope that Quantization
and Language Pair specialization can yield substantial model
compression.

\section*{Acknowledgments}
This work was generously supported by \href{https://cdac.in/}{C-DAC} via access to \href{https://cdac.in/index.aspx?id=hpc_nsf_siddhi-AI}{C-DAC Param Siddhi AI} HPC Cluster.

\FloatBarrier
\bibliographystyle{unsrt}
\bibliography{references}  

\end{document}